\documentclass{article}

\usepackage{microtype}
\usepackage{graphicx}
\usepackage{subfigure}
\usepackage{booktabs} 

\usepackage{hyperref}

\usepackage{algpseudocode}

\usepackage[accepted]{arxiv}

\usepackage{amsmath}
\usepackage{amssymb}
\usepackage{mathtools}
\usepackage{amsthm}

\usepackage[capitalize,noabbrev]{cleveref}

\theoremstyle{plain}

\theoremstyle{definition}

\theoremstyle{remark}

\usepackage[textsize=tiny]{todonotes}

\icmltitlerunning{Meta-Learning via Classifier(-free) Diffusion Guidance}

\begin{document}

\twocolumn[
\icmltitle{Meta-Learning via Classifier(-free) Diffusion Guidance}



\icmlsetsymbol{equal}{*}
\icmlsetsymbol{partlydone}{†}

\begin{icmlauthorlist}
\icmlauthor{Elvis Nava}{equal,aicenter,ini,srl}
\icmlauthor{Seijin Kobayashi}{equal,partlydone,ethcomp}
\icmlauthor{Yifei Yin}{ethcomp}
\icmlauthor{Robert K.~Katzschmann}{srl,aicenter}
\icmlauthor{Benjamin F.~Grewe}{ini,aicenter}
\end{icmlauthorlist}

\icmlaffiliation{aicenter}{ETH AI Center, ETH Z\"{u}rich, Z\"{u}rich, Switzerland}
\icmlaffiliation{ini}{Institute of Neuroinformatics, University of Z\"{u}rich and ETH Z\"{u}rich, Z\"{u}rich, Switzerland}
\icmlaffiliation{srl}{Soft Robotics Lab, ETH Z\"{u}rich, Z\"{u}rich, Switzerland}
\icmlaffiliation{ethcomp}{Dept.~of Computer Science, ETH Z\"{u}rich, Z\"{u}rich, Switzerland}

\icmlcorrespondingauthor{Elvis Nava}{enava@ethz.ch}
\icmlcorrespondingauthor{Robert K.~Katzschmann}{rkk@ethz.ch}
\icmlcorrespondingauthor{Benjamin F.~Grewe}{bgrewe@ethz.ch}

\icmlkeywords{machine learning, deep leaning, meta learning, hypernetworks, generative models, classifier guidance, contrastive learning, clip, classifier-free guidance, latent diffusion, diffusion models}

\vskip 0.3in
]



\printAffiliationsAndNotice{\icmlEqualContribution}{\partlydone} 

\begin{abstract}
We introduce meta-learning algorithms that perform zero-shot weight-space adaptation of neural network models to unseen tasks. Our methods repurpose the popular generative image synthesis techniques of natural language guidance and diffusion models to generate neural network weights adapted for tasks. We first train an unconditional generative hypernetwork model to produce neural network weights; then we train a second ``guidance'' model that, given a natural language task description, traverses the hypernetwork latent space to find high-performance task-adapted weights in a zero-shot manner. We explore two alternative approaches for latent space guidance: ``HyperCLIP''-based classifier guidance and a conditional Hypernetwork Latent Diffusion Model (``HyperLDM''), which we show to benefit from the classifier-free guidance technique common in image generation. Finally, we demonstrate that our approaches outperform existing multi-task and meta-learning methods in a series of zero-shot learning experiments on our Meta-VQA dataset. 
\end{abstract}

\section{Introduction}

\begin{figure}[h]
\includegraphics[width=\linewidth]{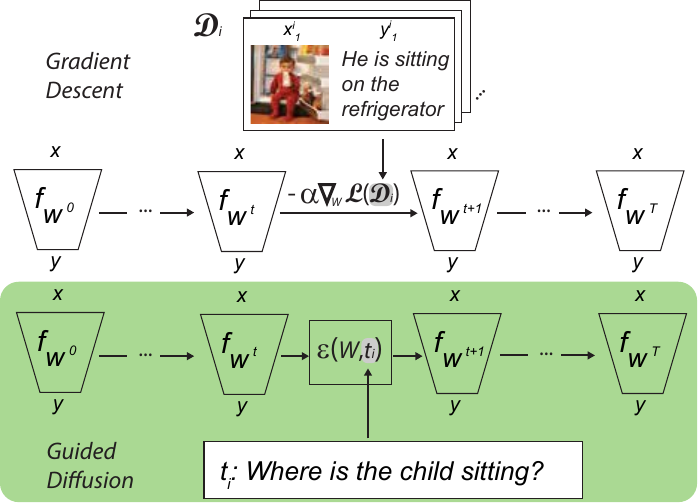}
\vspace{-0.2cm}
\caption{Given a task $\mathcal{T}_i$ and a network initialization $W^0$, with few-shot task data $\mathcal{D}_i$ one can use traditional gradient descent to perform task adaptation and obtain fine-tuned weights $W^T$. Our methods (green) instead do not require few-shot data, but use natural language descriptors $t_i$ to generate a surrogate adaptation towards $W^T$ in a zero-shot manner.}
\vspace{-0.2cm}
\label{fig:optimization}
\end{figure}

State-of-the-art machine learning algorithms often lack the ability to generalize in a sample efficient manner to new unseen tasks. In contrast, humans show remarkable capabilities in leveraging previous knowledge for learning a new task from just a few examples. Often, not even a single example is needed, as all relevant task information can be conveyed in the form of natural language instructions. Indeed, humans can solve novel tasks when prompted from a variety of different interaction modalities such as visual task observations or natural language prompts. In this work we present new meta-learning techniques that allow models to perform a similar kind of multi-modal task inference and adaptation in the weight-space of neural network models. In particular, we present two different approaches (\textbf{HyperCLIP guidance} and \textbf{HyperLDM}) that utilize natural language task descriptors for zero-shot task adaptation.

The development of deep learning models that perform quick adaptation to unseen tasks is the focus of the field of meta-learning. Meta-learning can be defined as a bi-level optimization problem, a trend stemming from the success of Model-Agnostic Meta-Learning \citep[MAML]{finn_model-agnostic_2017}: an outer loop meta-model is trained with the goal of improving the few-shot performance of a base model when fine-tuned on a variety of related tasks. MAML was specifically introduced as a gradient-based method to find a network initialization with high few-shot performance over an entire set of tasks. Recent progress in large scale transformer networks is however challenging this explicit meta-learning framework grounded in optimization over model weights. Large models trained on huge, rich, and diverse data sets have been shown to possess surprisingly good few-shot learning capabilities through in-context learning \citep{brown_language_2020}. Moreover, large scale pre-training and fine-tuning often outperforms explicit meta-learning procedures~\citep{mandi_effectiveness_2022}. \citet{brown_language_2020} dispense of the bi-level optimization formulation and use the word ``Meta-Learning'' to generally describe problem settings with an inner-loop/outer-loop structure, and use the words ``zero-shot'', ``one-shot'', or ``few-shot'' depending on how many demonstrations are provided in-context at inference time \footnote{See the footnote in page 4 of \citet{brown_language_2020}.}.

\begin{figure}[h]
\includegraphics[width=\linewidth]{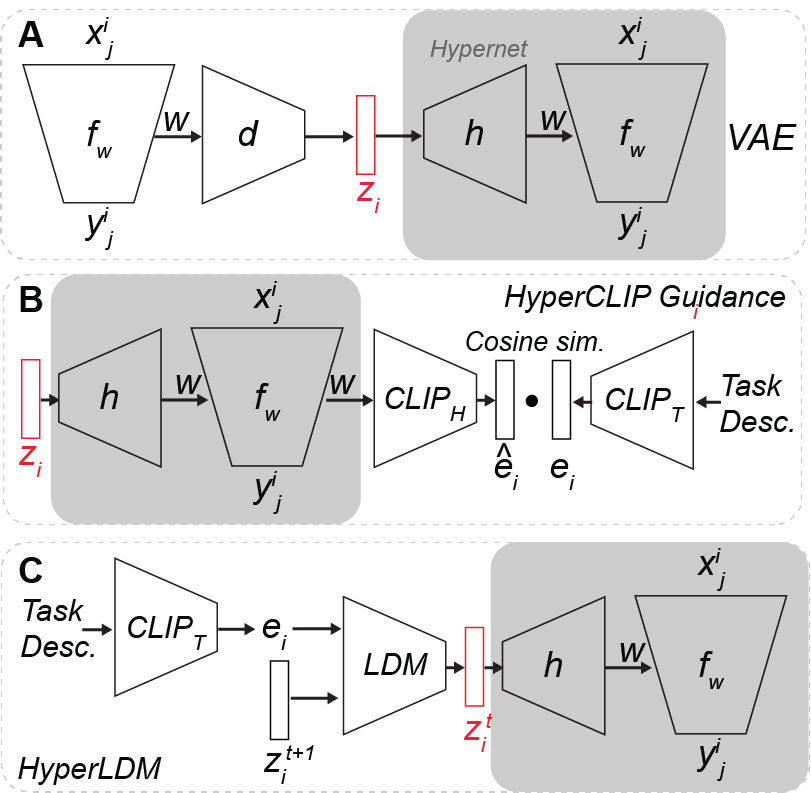}
\caption{Schematic of the three main components of our proposed meta-learning approach. \textbf{A.} A task-unconditional variational autoencoder (VAE) models the latent space of adapted weights $W$ for the network $f$ trained on data $(x^i_j,y^i_j)$. Its generator hypernetwork $h$ producing the weights (highlighted in gray) can be re-used in the task-conditional setting with our guidance techniques. \textbf{B.} Our HyperCLIP encoder $\text{CLIP}_H$ is contrastively trained to map network weights $W$ to the space of CLIP embeddings $e_i$. Then, given a new task with descriptor $t_i$, we can use CLIP guidance to find a VAE latent vector $z_i$ with embedding $e^{(H)}_i$ that has a high cosine similarity to a given task embedding $e^{(T)}_i$. \textbf{C.} Alternatively, our Hypernetwork Latent Diffusion Model (HyperLDM) learns, conditional on the task embedding $e_i$, to iteratively denoise a VAE latent vector $z_i^T,\ldots,z_i^0$ over $T$ iterations.}
\label{fig:schematic}
\end{figure}

These developments in transformer networks prompted us to develop alternative methods for meta-learning in weight-space which natively benefit from rich and multi-modal data like in-context learning. Inspired by recent advances in conditional image generation \citep{ramesh_hierarchical_2022, rombach_high-resolution_2022}, we recast meta-learning as a multi-modal generative modeling problem such that, given a task, few-shot data and natural language descriptions are considered equivalent conditioning modalities for adaptation (Figure~\ref{fig:optimization}). What we show is that popular techniques for the image domain, such as CLIP-based guidance \citep{gal_stylegan-nada_2021, patashnik_styleclip_2021}, denoising diffusion models \citep{ho_denoising_2020}, and classifier-free guidance \citep{dhariwal_diffusion_2021, ho_classifier-free_2021, nichol_glide_2022} can be repurposed for the meta-learning setting to generate adapted neural network weights. Using multi-step adaptation instead of traditionally conditioning the model on the natural language task information allows our models to achieve higher performance on each task by breaking down computations into multiple steps.

We approach the generation of neural network weights in two separate phases. In the \textit{unconditional pre-training} phase, we train a generative hypernetwork \citep{ha_hypernetworks_2016, schurholt_hyper-representations_2022} to map from its latent space to the weight space of a base model (Figure~\ref{fig:schematic}.A). In the \textit{guidance} phase, we learn language-conditioned models that can be used to traverse the hypernetwork latent space and find zero-shot adapted weights with high performance on our task (Figure~\ref{fig:schematic}.B and \ref{fig:schematic}.C). Our methods can thus benefit from large scale data through the pre-training phase, even when natural language descriptions are not available for all tasks.

We summarise our contributions as follows:

\textbf{1)} We introduce \textbf{HyperCLIP}, a contrastive learning method equivalent to Contrastive Language-Image Pre-training (CLIP)~\citep{radford_learning_2021}, producing CLIP embeddings of fine-tuned neural network weights. Using HyperCLIP as a guidance model then allows us to find task-adapted networks in the latent space of a hypernetwork model (Figure \ref{fig:schematic}.B).

\textbf{2)} We introduce Hypernetwork Latent Diffusion Models (\textbf{HyperLDM}) as a costlier but more powerful alternative to pure HyperCLIP guidance to find task-adapted networks within the latent space of a hypernetwork model (Figure \ref{fig:schematic}.C). We show how combining this approach with classifier-free guidance~\citep{ho_classifier-free_2021} improves the performance of generated base networks.

\textbf{3)} We demonstrate the usefulness of our methods on \textbf{Meta-VQA}, our modification of the \textit{VQA v2.0} dataset~\citep{goyal_making_2017} built to reflect the multi-task setting with natural language task descriptors. We show how our guidance methods outperform traditional multi-task and meta-learning techniques for zero-shot learning on this dataset.

\section{Meta-Learning with Multi-Modal Task Embeddings}
The setting we investigate is similar to the classic meta-learning framework, where we operate within a distribution of tasks $\mathcal{T}_i \sim p(\mathcal{T})$, each associated with a loss function $\mathcal{L}_{\mathcal{T}_i}$. Using a set of training tasks drawn from this distribution, our goal is to train a model such that it generally performs well on new unseen tasks drawn from $p(\mathcal{T})$. 

\subsection{Background on Model-Agnostic Meta-Learning} \label{sec:maml}
 In \citet{zintgraf_fast_2019}'s version of MAML, a model $g$ is composed of context parameters $\phi$ that are adapted on individual tasks, and shared parameters $\theta$ that are meta-trained and shared across tasks. MAML and its variants focus on the few-shot setting, which aims to learn an initialization for these parameters such that the model $g(\cdot, \theta, \phi)$ generalizes well on new tasks after fine-tuning $\phi$ on a few data points from that task. To train such a model, we sample training data $D_i$ from each task $\mathcal{T}_i$ and split it into a support set $D_i^{\text{s}}$ and a query set $D_i^{\text{q}}$. The MAML objective aims to optimize the validation score evaluated on the query set when fine-tuning $\phi$ on the support set:
\begin{equation}\label{maml}
    \min_{\theta, \phi} \mathop{\mathbb{E}}_{\mathcal{T}_i} \left[ \frac{1}{|D_i^{\text{q}}|}\sum_{(x,y)\in D_i^{\text{q}}}\mathcal{L}_{\mathcal{T}_i}\left(g(x, \theta, \mathcal{A}_{\mathcal{T}_i}(D_i^{\text{s}},\theta, \phi)), y\right) \right] \text{,} 
\end{equation}
where $\mathcal{A}_{\mathcal{T}_i}$ is some differentiable algorithm, typically implementing a variant of few-step gradient descent on the loss computed on the support set, \textit{e.g.}, in the case of one-step gradient descent:
\begin{equation}
    \mathcal{A}_{\mathcal{T}_i}(D_i^{\text{s}},\theta, \phi) = \phi -\eta \frac{1}{|D_i^{\text{s}}|} \sum_{(x',y')\in D_i^{\text{s}}} \nabla_\phi\mathcal{L}_{\mathcal{T}_i}(g(x',\theta, \phi),y')
\end{equation}
with some learning rate $\eta$.
The objective from Eq.~\ref{maml} is itself solved with gradient descent, by iteratively optimizing the parameters $\phi$ in the inner loop on the support set of a sampled task, and updating $\theta$ and the initialization of $\phi$ with their gradient with respect to the entire inner loop training process, averaged over batches of tasks.

\subsection{Natural Language Task Embeddings}

At test time, MAML-based meta-learning requires few-shot data $D_i^{\text{s}}$ from test tasks to adapt its unconditioned network parameters through gradient descent. In contrast, in this work, to perform zero-shot task adaptation, we utilize an additional high-level context embedding $e_i$ for each task $\mathcal{T}_i$. In practice, such embeddings can come from a natural language description $t_i$ of the task, which can be encoded into a task embedding using pre-trained language models.

A simple baseline for incorporating task embeddings into a model during training is by augmenting the input of the network, concatenating such input with the task embedding during the forward pass, or using custom conditioning layers such as FiLM \citep{perez_film_2017}. We instead consider the use of hypernetworks \citep{ha_hypernetworks_2016, zhao_meta-learning_2020}, a network that generates the weights of another network given a conditioning input.

\subsection{Meta-Learning and Hypernetworks}\label{sec:meta-learning-hypernetworks}

Given a neural network $f(\cdot,W)$ parametrized by a weight vector $W$, we reparametrize the model by introducing a hypernetwork $h$. The hypernetwork $h$ is parametrized by $\theta$, and generates weights $h(z, \theta)=W$ from an embedding $z$. The overall model is then defined as $f(\cdot, h(z, \theta))$. In the multi-task setting one can use a ``task-conditioned'' hypernetwork, in which the input embedding $z$ directly depends on the task $\mathcal{T}_i$ (\textit{e.g.}~$z=e_i$). In this work, we will also consider ``unconditional'' hypernetworks, trained as generative models (see Section \ref{sec:hypernet-generative}), with input embeddings $z$ that don't depend on the task, but may for example be normally distributed.

Before introducing our new zero-shot techniques, we construct a hypernetwork-based baseline by rewriting the MAML objective (Eq.~\ref{maml}) with respect to the hypernetwork weight $\theta$ as
\begin{equation}\label{maml-hnet}
    \min_{\theta} \mathop{\mathbb{E}}_{\mathcal{T}_i} \left[ \frac{1}{|D_i^{\text{q}}|}\sum_{(x,y)\in D_i^{\text{q}}}\mathcal{L}_{\mathcal{T}_i}\left(f(x,h( \mathcal{A}_{\mathcal{T}_i}(D_i^{\text{s}},\theta,z), \theta))), y\right) \right] \text{.} 
\end{equation}
 Forcing $\mathcal{A}_{\mathcal{T}_i}(D_i^{\text{s}}, \theta, z)=e_i$, we recover the simple multi-task objective of a task-conditioned hypernetwork optimizing for zero-shot performance, taking $e_i$ directly as input. When $\mathcal{A}_{\mathcal{T}_i}$ is instead the gradient descent algorithm on $z$, the objective corresponds to a variant of MAML, optimizing the few-shot performance of $h$ when only adapting the embedding in the inner loop, initialized at $z$. For more details related to the baselines, see Appendix \ref{appendix:method-details}.
 
\section{Hypernetworks as Generative Models}\label{sec:hypernet-generative}

A rich literature exists on hypernetworks interpreted as generative models of base network weights (see Section \ref{section:related-work}). Our work builds upon this interpretation to adapt multi-modal generative modeling techniques to the meta-learning domain.

In generative modeling, we aim to learn the distribution $p(x)$ over a high dimensional data domain $\mathcal{X}$, such as images, given samples from the distribution. To do so, we resort to techniques such as variational inference, adversarial training, or diffusion models. Our meta-learning setting can analogously be framed as modeling a distribution of diverse high-dimensional base network weights $W$. In the Bayesian setting, this distribution is made explicit as we seek to model the posterior $p(W|D)$ given data $D$, but the framework is still useful even when no explicit posterior distribution is assumed, as it is the case for deep ensembles. In the present work, we indeed avoid explicit Bayesian inference: for each training task $\mathcal{T}_i$, we fine-tune the base model $f(x,W) = y$ on it, and use the resulting $W_i$ as a training sample to train a generative model of network weights.

The fundamental building block of our unconditional generative model is the hypernetwork $h(z,\theta) = W$ that we can train in two ways:
\textbf{1. \textbf{HVAE}:} We define a Hypernetwork VAE (Figure \ref{fig:schematic}.A), which, given samples of fine-tuned base network weights $W_i$, learns a low-dimensional normally distributed latent representation $z_i$. The encoder $d(W_i, \omega) = (\mu_{z_i}, \Sigma_{z_i})$ with parameters $\omega$ maps base network weights to means and variances used to sample a latent vector $z_i$, while the decoder (or generator) is a classic hypernetwork $h(z_i, \theta) = W_i$ which reconstructs the network weights from the latent vector (See Appendix \ref{appendix:generative-hypernet}). This VAE setup is analogous to that proposed in recent work on \textit{hyper-representations} \citep{schurholt_hyper-representations_2022}.
\textbf{2. \textbf{HNet}:} Using MAML, we learn both an \textit{initialization embedding} $z$ and hypernetwork weights $\theta$ such that, when fine-tuning only the embedding $z$ on each task $\mathcal{T}_i$, we obtain high-performing base networks with weights $W_i = h(z_i,\theta)$. Concretely, we optimize $\theta$ and the initialization of $z$ following the objective in Eq.~\ref{maml-hnet} (see Section \ref{sec:meta-learning-hypernetworks}).

Up to this point, we trained an unconditional hypernetwork generative model of neural network weights, comprising the \textit{unconditional pre-training} phase of our meta-learning approach. This gives us a powerful generator $h(z, \theta) = W$, which maps from its latent space to the weight-space of our base network. In the next Section, we investigate how to then perform task-conditional \textit{guidance} within this latent space, finding adapted latent embeddings $z_i$ for our test tasks in a zero-shot manner.

\section{HyperCLIP: Training a CLIP Encoder for the ``Model-Parameters Modality''}

To define the first of our two meta-learning \textit{guidance} techniques, we borrow from the field of multi-modal contrastive learning. More specifically, we build on top of Contrastive Language-Image Pre-training (CLIP) \citep{radford_learning_2021}, a popular method for joint learning of language and image embeddings with applications to zero-shot and few-shot classification.

In the original CLIP formulation, separate text and image encoders are trained such that, given a bi-modal sample $(x_i, t_i)$ of an image and its corresponding language caption, their representations ($\text{CLIP}_I(x_i) = e^{(I)}_i$ and $\text{CLIP}_T(t_i) = e^{(T)}_i$) are aligned across modalities. Specifically, the formulation maximizes the cosine similarity $e^{(I)\top}_i e^{(T)}_j / \|e^{(I)}_i\| \|e^{(T)}_j\|$ for pair-wise matches ($i=j$) and minimizes the cosine similarity for non-matches ($i\neq j$). Beyond the original language-image setting, the CLIP approach can be easily adapted to include additional modalities, aligning the representation of more than two encoders at a time. Existing works such as AudioCLIP \citep{guzhov_audioclip_2022} demonstrate the possibility of training an encoder for an additional modality such as audio on the side of the pre-trained frozen CLIP language-image encoders.

\subsection{Contrastive Learning on Neural Network Weights}

\begin{figure*}[h]
\begin{center}
\includegraphics[width=\linewidth]{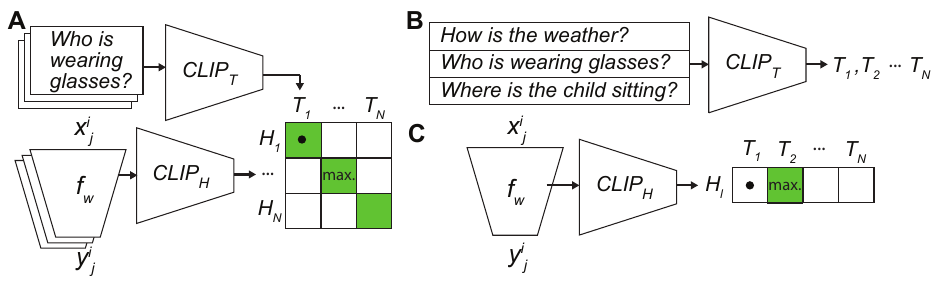}
\vspace{-5mm}
\end{center}
\caption{Our HyperCLIP encoder $\text{CLIP}_H$ is contrastively trained to map neural network weights $W$ to the latent space of a pre-trained language encoder $\text{CLIP}_T$, which we use to embed the natural language questions associated to the tasks (see \textbf{A}). To perform task inference given an already fine-tuned network, one can encode all candidate task questions using the language CLIP encoder (see \textbf{B}), then encode the fine-tuned network weights with HyperCLIP (see \textbf{C}), and finally infer the correct task with a softmax operation over cosine similarities between HyperCLIP and language CLIP embeddings.}
\label{fig:hyperclip-schematic}
\end{figure*}

In our work, we consider multi-modal representations of meta-learning tasks $\mathcal{T}_i$. A descriptor of a task may come from the language modality ($t_i$), but potentially also from image, video or audio modalities. When we fine-tune a base machine learning model $f(x,W_i) = y$ for task $\mathcal{T}_i$, we then also consider the fine-tuned base model weights $W_i$ as being part of an alternative \textit{model-parameters modality} that describes task $\mathcal{T}_i$. Fine-tuned network weights from the \textit{model-parameters modality} can then be paired in contrastive learning with the other multi-modal descriptions of $\mathcal{T}_i$. 
We thus define our new \textbf{HyperCLIP} encoder $\textit{CLIP}_H(W_i) = e^{(H)}_i$, taking fine-tuned neural network weights $W_i$ as input, and outputting a CLIP embedding $e^{(H)}_i$ optimized for high cosine similarity with the CLIP embedding for the textual (or image, audio, etc.) descriptor of the task. Figure \ref{fig:hyperclip-schematic} and to Algorithm \ref{algo:hyperclip} illustrate the approach.

\begin{algorithm}[h]
\caption{HyperCLIP Training}\label{algo:hyperclip}
\begin{algorithmic}
\State sample a batch of tasks $\mathcal{T}_{i=1,\ldots,N}$ with loss functions $\mathcal{L}_{\mathcal{T}_i}$, training data $D_i^{\text{train}}$ and text $t_i$
\State define two $N$-sized arrays of $d$-dimensional embeddings $T \in \mathbb{R}^{N\times d}$ and $H \in \mathbb{R}^{N\times d}$
\For{$i=1,\ldots,N$}
	\State $T[i] = \text{CLIP}_T(t_i) \; / \; \|\text{CLIP}_T(t_i)\|$
	\State Fine-tune $W_i$: $\min_{W}  \sum_{(x',y')\in D_i^{\text{train}}}\mathcal{L}_{\mathcal{T}_i}(f(x', W), y')$
	\State $H[i] = \text{CLIP}_H(W_i) \; / \; \|\text{CLIP}_H(W_i)\|$
\EndFor
\State $\text{loss} = \left(\mathcal{L}_\text{cross-entropy}(TH^\top) + \mathcal{L}_\text{cross-entropy}(HT^\top)\right) / \; 2$
\State Update weights of $\text{CLIP}_H(.)$ using $\nabla \text{loss}$
\end{algorithmic}
\end{algorithm}

\subsection{Classifier-Guided Meta-Learning}\label{section:classguidance}

On their own, CLIP encoders are not capable of data generation. Recent popular image synthesis techniques, however, use CLIP encoders or other classifiers to \textit{guide} generation from pre-trained unconditional generative models. \textit{Classifier guidance} or \textit{CLIP guidance} \citep{gal_stylegan-nada_2021, patashnik_styleclip_2021} use gradients with respect to a classifier or CLIP encoder to traverse a generative model's latent space.

In this work, we introduce \textbf{HyperCLIP guidance}, the first algorithm for classifier guidance in the meta-learning setting (Figure \ref{fig:schematic}.B). Given a previously unseen validation task $\mathcal{T}_i$ and an unconditional generative hypernetwork model $h(z, \theta) = W$, we use a trained HyperCLIP encoder $\text{CLIP}_H(W) = e^{(H)}$ to guide the exploration of the hypernetwork's latent space and find a set of base weights $W_i$ with high zero-shot performance for $\mathcal{T}_i$. Specifically, as long as we are given a starting hypernetwork latent vector $z^0$ and a textual description $t_i$ of the task, we can update $z^0$ with gradient descent over the guidance loss
\begin{equation}\label{eq:hyperclip-guidance}
    \mathcal{L}_\text{g}(z) = - \frac{\text{CLIP}_H\left( h(z, \theta) \right)^\top \text{CLIP}_T(t_i)}{\|\text{CLIP}_H\left( h(z, \theta) \right)\| \|\text{CLIP}_T(t_i)\|} + \lambda \| z - z^0 \| \text{,}
\end{equation}
and then run the optimized latent vectors $\hat{z}_i$ through the generative hypernetwork to find adapted zero-shot base network weights $h(\hat{z}_i, \theta) = \hat{W}_i$ that perform well for the task.

\section{HyperLDM: Task-conditional Diffusion of Hypernetwork Latents}

Rapid innovation in the image synthesis community recently led to simple CLIP-guidance being largely overcome in favor of applying classifier guidance and classifier-free guidance during the sampling process of a Diffusion Model \citep{dhariwal_diffusion_2021, ho_classifier-free_2021, kim_guided-tts_2022-1, crowson_v-diffusion_2022, nichol_glide_2022, rombach_high-resolution_2022}. To investigate whether these advances also apply for our meta-learning setting, we introduce \textbf{HyperLDM}, a diffusion-based technique as an alternative to the previously introduced HyperCLIP guidance.

\subsection{(Latent) Diffusion Models}

Denoising Diffusion Probabilistic Models \citep[DDPM]{sohl-dickstein_deep_2015, ho_denoising_2020} are a powerful class of generative models designed to learn a data distribution $p(x)$. They do so by learning the inverse of a \textit{forward diffusion process} in which samples $x^0$ of the data distribution are slowly corrupted with additive Gaussian noise over $T$ steps with a variance schedule $\beta_1,\ldots,\beta_T$, resulting in the Markov Chain
\begin{align}
    q(x^t|x^{t-1}) &= \mathcal{N}(x^t; \sqrt{1-\beta_t}x^{t-1}, \beta_t \mathbf{I})\\ q(x^{1:T}|x^0) &= \prod_{t=1}^T{q(x^t|x^{t-1})}\text{.}
\end{align}

A property of such a process is that we can directly sample each intermediate step from $x^0$ as $x^{t} = \sqrt{\Bar{\alpha}_t} x^0 + (\sqrt{1-\Bar{\alpha}_t}) \epsilon$ given $\epsilon \sim \mathcal{N}(0,\mathbf{I})$, $\alpha_t = 1 - \beta_t$ and $\Bar{\alpha}_t = \prod_{s=1}^t\alpha_t$. Then, to learn the reverse process $p_\psi(x^{t-1}|x^t)$, we parametrize the timestep-dependent noise function $\epsilon_\psi(x^t,t)$ with a neural network and learn it by optimizing a simplified version of the variational lower bound on $p(x)$
\begin{equation}
    \mathcal{L}_{\text{DM}}(\psi) = \mathbb{E}_{x,\epsilon\sim\mathcal{N}(0,1),t}\left[ \| \epsilon - \epsilon_\psi(x^t, t) \|^2_2 \right] \text{.}
\end{equation}
Sampling from the reverse process can then be done with
\begin{equation}
    x^{t-1} = \frac{1}{\sqrt{\alpha_t}} \left( x^t - \frac{\beta_t}{\sqrt{1 - \Bar{\alpha}_t}} \epsilon_\psi(x^t, t) \right) + \sigma_t\xi \text{,}
\end{equation}
with $\mathbf{\xi} \sim \mathcal{N}(0,\mathbf{I})$ and $\sigma_t$ chosen between $\beta_t$ and $\beta_t / \sqrt{1 - \Bar{\alpha}_t}$. Sampling from the learned diffusion model can be seen as analogue to Langevin Dynamics, a connection explicitly made in works exploring the diffusion technique from the ``score matching'' perspective \citep{song_generative_2019, song_score-based_2020}.

In our meta-learning setting, we aim to train a diffusion model which generates adapted zero-shot base network weights $\hat{W}_i$ that perform well for task $\mathcal{T}_i$. Thus, our diffusion model has to be conditional on a task embedding $e_i$. Moreover, in order to speed up training and leverage our previously trained generative hypernetwork $h(z, \theta)$, we define the diffusion process on latent vectors instead of doing so in weight space, emulating the Latent Diffusion technique from \citet{rombach_high-resolution_2022}.

We then propose a Hypernetwork Latent Diffusion Model (\textbf{HyperLDM}), which learns to sample from the conditional distribution of fine-tuned latent vectors $p(z^0|e_i)$ given a language CLIP embedding corresponding to the task. The HyperLDM neural network fits the noise function $\epsilon_\psi(z^t,t,e_i)$, and is learned by optimizing the reweighted variational lower bound, which in this setting is
\begin{equation}\label{eq:hyperldm-loss}
    \mathcal{L}_{\text{LDM}}(\psi) = \mathbb{E}_{\mathcal{T}_i,d(W_i),\epsilon\sim\mathcal{N}(0,1),t}\left[ \| \epsilon - \epsilon_\psi(z^t,t,e_i) \|^2_2 \right] \text{.}
\end{equation}

\subsection{Classifier-Free Guidance for Meta-Learning} \label{section:class-free}

To improve the quality of sampled networks, the classifier guidance technique presented in Section \ref{section:classguidance} can be also combined together with diffusion models. The gradient of an auxiliary classifier (or CLIP encoder) can be added during sampling to induce an effect similar to GAN truncation \citep{brock_large_2018}, producing samples that are less diverse but of higher quality.

The classifier-free guidance technique \citep{ho_classifier-free_2021, nichol_glide_2022} allows us to leverage a conditional diffusion model to obtain the same effect as above, without the auxiliary classifier. To do so, we train the conditional network $\epsilon_\psi(z^t,t,e_i)$ to also model the unconditional case $\epsilon_\psi(z^t,t)$. One way of doing this is with \textit{conditioning dropout}, simply dropping the conditional input $e_i$ for a certain percentage of training samples, inputting zeros instead. We can then sample at each diffusion iteration with
\begin{equation}
    \tilde{\epsilon}_\psi(z^t,t,e_i) = \left(1 - \gamma\right) \epsilon_\psi(z^t,t,0) + \gamma \epsilon_\psi(z^t,t,e_i) \text{.}
\end{equation}
For $\gamma = 0$, this recovers the unconditional diffusion model, while for $\gamma = 1$ it recovers the standard task-conditional model. For $\gamma > 1$, we instead obtain the classifier-free guidance effect, which we show results in the sampling of latent vectors $\hat{z}_i$ corresponding to higher-performing task-conditional network weights $h(\hat{z}_i, \psi) = \hat{W}_i$. We point to a more in-depth discussion on classifier-free guidance and its connection to score matching in Appendix \ref{appendix:class-free}.

\section{Experimental Setup and Results}

In this section, we demonstrate the competitiveness of our two approaches in zero-shot image classification experiments against a series of traditional meta-learning techniques. Throughout our experiments, we fix the choice of base network model $f$ to a CLIP-Adapter model (see Appendix \ref{appendix:architecture}), only varying the meta-learning techniques employed to obtain adapted base model weights. The CLIP-Adapter base model makes use of pre-trained CLIP encoders to obtain high base performance on image classification with textual labels, while maintaining a relatively small trainable parameter count. It should not be confused with the usage of CLIP encoders to produce task embeddings, or to train HyperCLIP, all of which happens at the meta-level.

\subsection{The Meta-VQA Dataset}

To evaluate the performance of our methods, we utilize a dataset that reflects the setting of meta-learning with multi-modal task descriptors. Existing meta-learning benchmarks such as MiniImagenet \citep{ravi_optimization_2016} or CIFAR-FS \citep{bertinetto_meta-learning_2018} are unsuitable, as they are built for the traditional few-shot learning setting, in which the task $\mathcal{T}_i$ is not associated with task descriptors but is meant to be inferred through exposure to the support set $D_i^{\text{s}}$. We thus introduce our own \textbf{Meta-VQA} dataset, a modification of the VQA v2.0 dataset~\citep{goyal_making_2017} for Visual-Question-Answering. The dataset is composed of training and test tasks $\mathcal{T}_i$, each associated with a natural language question $t_i$ and a mini image classification dataset $(x_j^i,y_j^i) \in D_i$. We refer to Appendix \ref{appendix:meta-vqa} for a more in-depth discussion.

\begin{figure}[h]
\begin{center}
\includegraphics[width=\linewidth]{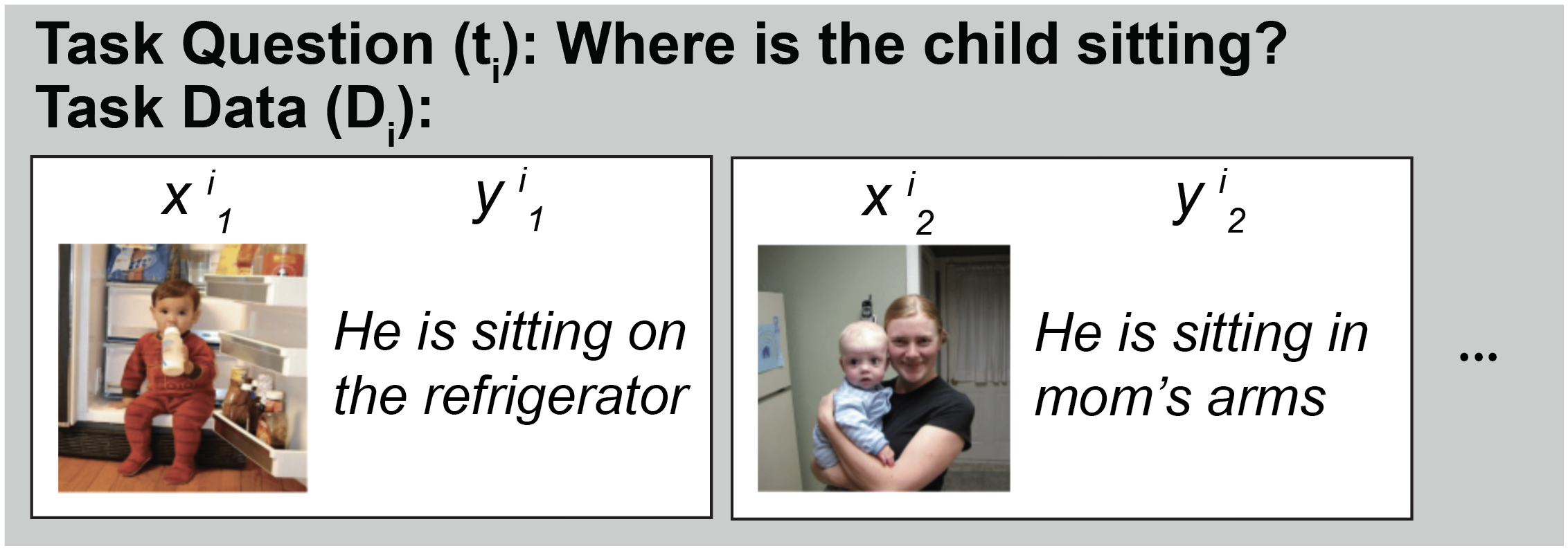}
\end{center}
\caption{Example classification task from Meta-VQA, adapted from VQA v2 \citep{goyal_making_2017}. A task $\mathcal{T}_i$ is associated to a single question $t_i$ and multiple image-answer tuples $(x_j^i,y_j^i)$.}
\end{figure}

\subsection{Zero-Shot Task Adaptation Experiments} \label{sec:experiment-results}

\begin{table*}[ht]
\caption{Zero-shot accuracy (mean $\pm$ s.d.) averaged over Meta-VQA test tasks. Results should be interpreted as relative to a performance ceiling of 60.24\% obtainable when task data is available (few-shot learning), and with our fixed choice of base model (see Appendix \ref{appendix:fewshot}). The columns separate the setting in which only half of task descriptors/questions are given (50\% Q.), and that in which all of the task descriptors are given (100\% Q.). (* ours)}
\label{tab:zshot-fshot}
\begin{center}
\begin{tabular}{l l l}
\toprule
\multicolumn{1}{c}{\bf Method}  &
\multicolumn{1}{c}{\bf Zero-shot (50\% Q.)} &
\multicolumn{1}{c}{\bf Zero-Shot (100\% Q.)}
\\ \midrule
CLIP as Base Model & \multicolumn{2}{c}{44.99} \\
\midrule
Uncond.~Multitask   & \multicolumn{2}{c}{53.75 ($\pm$ 0.36)} \\
Uncond.~MNet-MAML   & \multicolumn{2}{c}{53.04 ($\pm$ 0.69)} \\
Uncond.~MNet-FOMAML & \multicolumn{2}{c}{53.04 ($\pm$ 0.42)} \\
Uncond.~HNet-MAML   & \multicolumn{2}{c}{53.37 ($\pm$ 0.29)} \\
\midrule
Cond.~Multitask     & 51.68 ($\pm$ 0.33) & 54.12 ($\pm$ 0.80) \\
Cond.~Multitask FiLM          & 51.60 ($\pm$ 0.56) & 53.84 ($\pm $ 0.61)\\
Cond.~HNet-MAML     & 51.54 ($\pm$ 0.63) & 53.02 ($\pm$ 0.20) \\
\midrule
* HNet + HyperCLIP Guidance   & 53.51 ($\pm$ 0.22) & 53.98 ($\pm$ 0.54)\\ 
* HVAE + HyperCLIP Guidance & 53.82 ($\pm$ 0.07) & 53.91 ($\pm$ 0.08)\\
* HNet + HyperLDM $\gamma=1$ & 53.66 ($\pm$ 0.25) & 54.06 ($\pm$ 0.21) \\
* HNet + HyperLDM $\gamma=1.5$ & 54.08 ($\pm$ 0.11) & 54.30 ($\pm$ 0.27)\\
* HVAE + HyperLDM $\gamma=1$ & 54.72 ($\pm$ 0.23) & 55.03 ($\pm$ 0.32) \\
* HVAE + HyperLDM $\gamma=1.5$  & \textbf{54.84} ($\pm$ 0.24) & \textbf{55.10} ($\pm$ 0.08)\\
\bottomrule
\end{tabular}
\end{center}
\end{table*}

\begin{figure*}[htb]
\begin{center}
\subfigure{\includegraphics[width=0.49\linewidth]{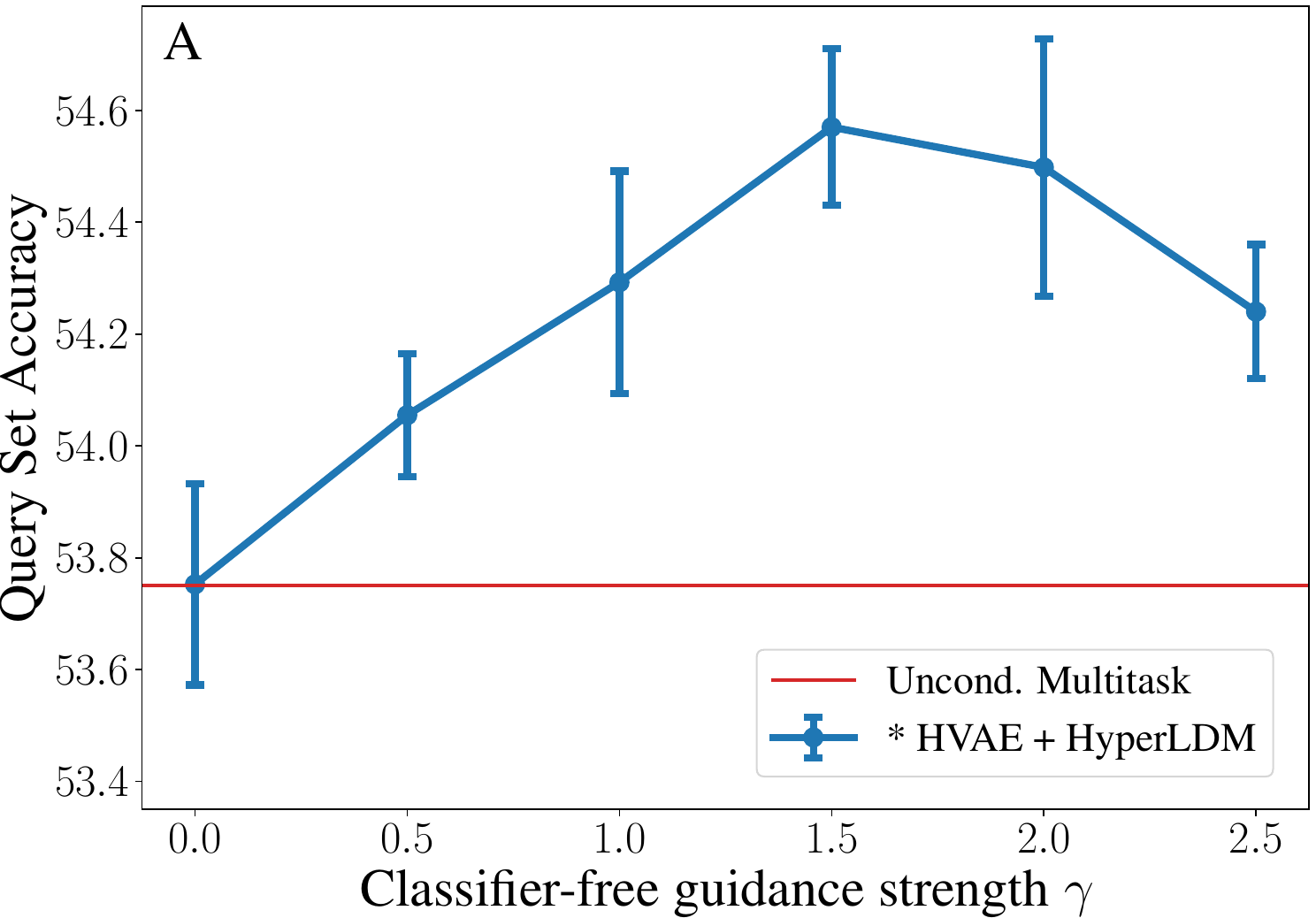}}
\hfill
\subfigure{\includegraphics[width=0.49\linewidth]{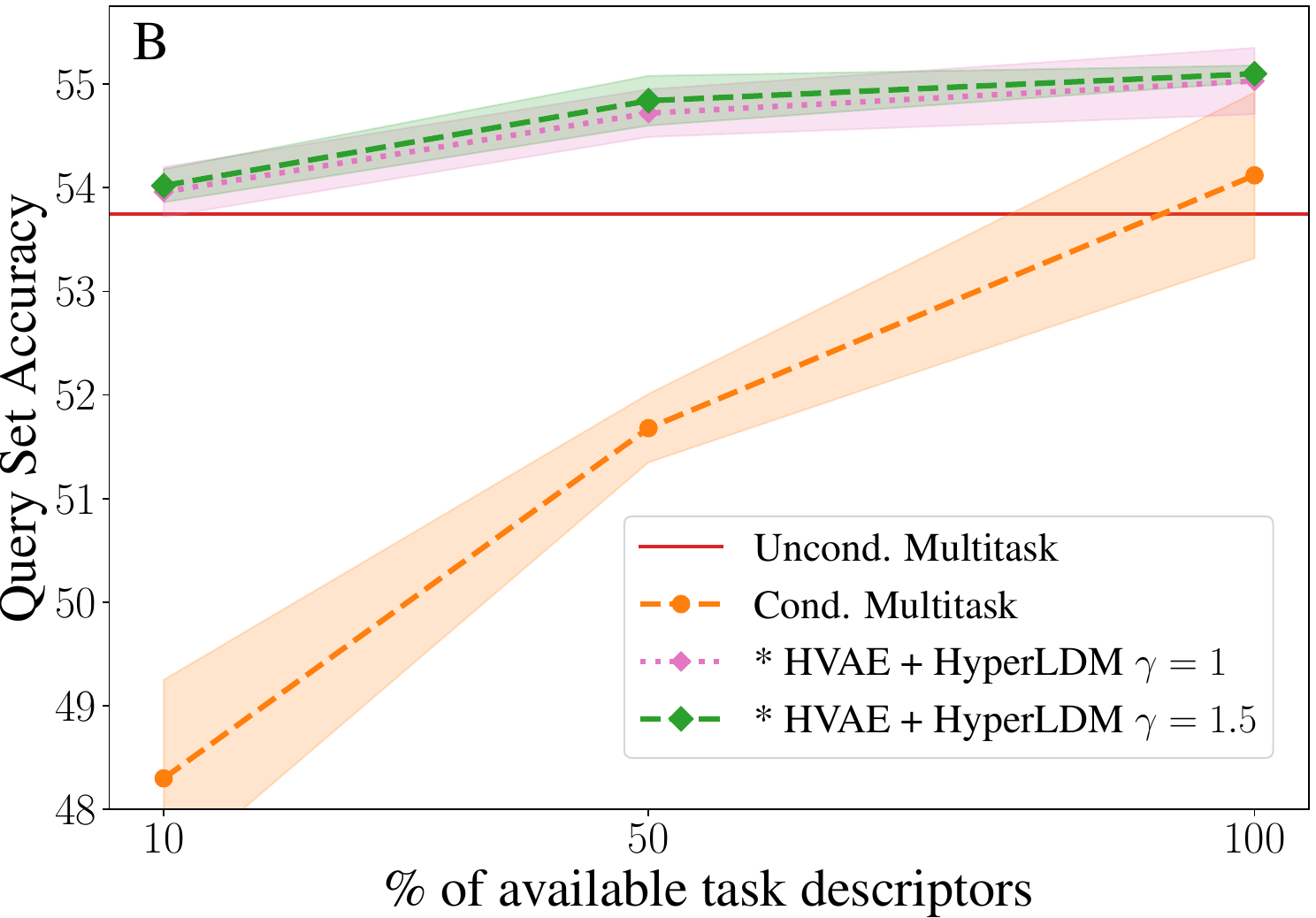}}
\end{center}
\caption{\textbf{A)} Zero-shot performance of HyperLDM (mean $\pm$ s.d.) over different classifier-free guidance parameters $\gamma$. For $\gamma=0$ we sample from an unconditional latent diffusion model. For $\gamma=1$ we sample with classic conditioning. For $\gamma>1$, we are in the classifier-free guidance regime. \textbf{B)} Zero-shot performance of HyperLDM (mean $\pm$ s.d.) against baselines in the setting where only a fraction of natural language task labels are given.}
\label{fig:experiments-1}
\end{figure*}

In Table \ref{tab:zshot-fshot} we show how our methods compare to a series of baselines when tested on the Meta-VQA dataset in the zero-shot setting. For each training task $\mathcal{T}_i$, the algorithms are given access to the full support and query sets $D_i^{\text{s}}$, $D_i^{\text{q}}$, together with the question (task descriptor) $t_i$. At test time, in the zero-shot setting, only the task descriptors $t_i$ are given, and the algorithms are tasked with predicting the correct labels of images in the query set $D_i^{\text{q}}$.

In addition, we also simulate a setting in which we possess a larger ``unconditional'' pre-training dataset. Our two-phased approach, which separates generative model pre-training and guidance, benefits from unconditional data: tasks without language descriptors can still be used to learn the unconditional HNet/HVAE model. To test this, we conduct additional runs in which we train our model while only keeping a fraction of task descriptors from the Meta-VQA dataset.

Using the original CLIP for zero-shot image classification (\textbf{CLIP as Base Model}) provides a 44.99\% \textit{floor} for performance on Meta-VQA. All other techniques will use CLIP-Adapter as base model, as previously mentioned. We also obtained an approximate 60.24\% performance \textit{ceiling} from the best method in the few-shot setting, in which models have also access to a data support set $D_i^{\text{s}}$ for every test task (see Appendix \ref{appendix:fewshot}). Our zero-shot techniques cannot surpass this ceiling while keeping the choice of base model fixed. The zero-shot scores should then be judged within a range between 44.99\% and 60.24\% accuracy.

We then benchmark several unconditional and conditional methods, with only conditional methods having access to language task descriptors. We apply MAML and its first-order variant FOMAML \citep{nichol_first-order_2018} directly to the base network (\textbf{MNet-MAML}, \textbf{MNet-FOMAML}), and to both an unconditional hypernetwork (\textbf{Uncond.~HNet-MAML}, as in Section \ref{sec:meta-learning-hypernetworks}) and a conditional one (\textbf{Cond.~HNet-MAML}). We also benchmark against standard multitask learning (\textbf{Uncond.~Multitask}, \textbf{Cond.~Multitask}) on the base model without hypernetworks, and conditional multitask learning with the classic FiLM layer \citep{perez_film_2017} (\textbf{Cond.~Multitask FiLM}). Note that the multitask approach, at least in this setting, leads to better zero-shot models than MAML, which instead optimizes for few-shot performance. We refer to Appendix \ref{appendix:architecture} and \ref{appendix:method-details} for more details on each algorithm.

We then test out two approaches, \textbf{HyperCLIP Guidance} and \textbf{HyperLDM}, when trained on top of either a hypernetwork or a VAE generator (Section \ref{sec:hypernet-generative}, see also Appendix \ref{appendix:architecture} and \ref{appendix:our-method-details} for more detail). \textbf{HyperCLIP Guidance} allows for faster sampling than \textbf{HyperLDM} but is generally less performant, still, it performs on par with or slighly improves upon all other zero-shot baselines except for \textbf{Cond.~Multitask}. The best performing model for the zero-shot setting is \textbf{HVAE + HyperLDM}, and specifically for classifier-free guidance with $\gamma=1.5$. As illustrated in Figure \ref{fig:experiments-1}.A, to further show the effectiveness of the classifier-free guidance technique, we run a different experiment sweeping over several candidate $\gamma$ parameters to find that the optimum occurs for $\gamma>1$. As shown in Figure \ref{fig:experiments-1}.B, when training our model while only keeping 50\% or 10\% of task descriptors, traditional \textbf{Cond.~Multitask} learning heavily overfits, while \textbf{HyperLDM} is almost not affected due to its two-phased training regime based on an uncondtional VAE. The gap between the multitask baseline and our HyperLDM technique is particularly striking in this setting.

\section{Related Work}\label{section:related-work}

\paragraph{Hypernetworks} 
By introducing multiplicative interactions within neural networks \citep{jayakumar_multiplicative_2019}, hypernetworks \citep{ha_hypernetworks_2016} have been shown to allow the modeling of diverse target network weights in, \textit{e.g.}, continual learning, even in the compressive regime \citep{von_oswald_neural_2021,von_oswald_continual_2020} without loss of performance. For a given supervised problem, hypernetworks have been used to model the complex Bayesian posterior of the weights in conjunction with variational inference \citep{henning_approximating_2018, krueger_bayesian_2018}. Similar approaches have been used for continual learning \citep{henning_posterior_2021}. Another vein of work consists in using hypernetworks to distill ensembles of diverse networks \citep{wang_adversarial_2018,ratzlaff_hypergan_2020,von_oswald_neural_2021}. Recent work also explored the properties of hypernetworks as autoencoder generative models of network weights \citep{schurholt_hyper-representations_2022}.

\paragraph{Meta learning} 
In the context of meta-learning, hypernetworks have been successfully used in combination with popular gradient-based meta-learning methods \citep{finn_model-agnostic_2017,zintgraf_fast_2019,zhao_meta-learning_2020,flennerhag_meta-learning_2020}. More generally, related works have shown the usefulness of learning a low dimensional manifold in which to perform task-specific gradient-based adaptation at meta test time \citep{rusu_meta-learning_2018,von_oswald_learning_2021,lee_gradient-based_2018}, instead of directly adapting in weight space. Recent works bypass the formal bi-level formulation of meta-learning \citep{brown_language_2020} by, \textit{e.g.}, using transformers to directly map the few-shot examples to the weights of the target network \citep{zhmoginov_hypertransformer_2022}.

\paragraph{Generative Modeling and Classifier(-free) guidance}
A plethora of techniques have been proposed for the generation of samples from high-dimensional domains such as images, such as Generative Adversarial Networks \citep[GANs]{goodfellow_generative_2014, brock_large_2018} and Variational Autoencoders \citep[VAEs]{kingma_auto-encoding_2014}. Denoising Diffusion Probabilistic Models \citep[DDPM]{sohl-dickstein_deep_2015, ho_denoising_2020} overcome common issues in generative modeling using a simple likelihood-based reconstruction loss for iterative denoising, and have been shown to achieve state-of-the-art results in high resolution image generation \citep{dhariwal_diffusion_2021, rombach_high-resolution_2022}. Several techniques have been proposed for effective conditional sampling in generative and diffusion models, such as classifier/CLIP guidance \citep{dhariwal_diffusion_2021, gal_stylegan-nada_2021, patashnik_styleclip_2021} and classifier-free guidance \citep{ho_classifier-free_2021,crowson_v-diffusion_2022,nichol_glide_2022}. Diffusion models with classifier-free guidance have also been successfully applied in non-visual domains, such as audio generation \citep{kim_guided-tts_2022-1} and robotic planning \citep{janner_planning_2022}.

\paragraph{Zero-shot learning} There exists a large literature on zero-shot learning, including both established benchmarks and well known methods \citep{han_contrastive_2021,su_distinguishing_2022,gupta_generative_2021}. While these zero-shot learning works consider the zero-shot performance on unseen class labels within a single classification task, our setting considers that of the zero-shot performance where test tasks themselves are unseen, thus raising the zero shot problem to the task-level.

\section{Conclusion}
In this work we introduced a framework that re-interprets meta-learning as a multi-modal generative modeling problem. Our HyperCLIP guidance and HyperLDM methods leverage this insight to generate task-adapted neural network weights in a zero-shot manner given natural language instructions, and constitute the first application of the CLIP guidance and classifier-free guidance techniques from image generation to the meta-learning domain. Our experiments show that our methods successfully make use of external task descriptors to produce high-performance adapted networks in the zero-shot setting.

\section*{Acknowledgments}
We are grateful for funding received by the ETH AI Center, Swiss National Science Foundation (B.F.G.~CRSII5-173721 and 315230 189251), ETH project funding (B.F.G.~ETH-20 19-01), the Human Frontiers Science Program (RGY0072/2019), and Credit Suisse Asset Management.

\subsection*{Code Release}
Our code is available at \url{https://github.com/elvisnava/hyperclip}.

\subsection*{Author Contributions}
\paragraph{Elvis Nava *} Original idea, implementation and experiments (guidance techniques and diffusion), paper writing (overall manuscript structure, generative modeling and guidance sections, experimental results section).
\paragraph{Seijin Kobayashi *} Implementation and experiments (MAML and hypernetworks, HyperCLIP guidance), paper writing (meta-learning background and related work).
\paragraph{Yifei Yin} Creation of the Meta-VQA dataset, initial experiments on the CLIP-Adapter base model.
\paragraph{Robert K.~Katzschmann} Conceptualising of the project idea (focus on future robotics applications), participating in project meetings, giving feedback and conceptually guiding the approaches, giving feedback to the manuscript.
\paragraph{Benjamin F.~Grewe} Conceptualisation of the project with a focus on using language task descriptors, senior project lead, participating in project meetings, providing conceptual input for developing the two approaches, giving feedback on the manuscript, developing intuitive schematics for Figure 1 and 2.


\bibliography{references}
\bibliographystyle{icml2023}

\newpage
\appendix
\onecolumn
\section{Appendix}

\subsection{Classifier-Free Guidance} \label{appendix:class-free}
We hereby provide a rationale for the use of classifier guidance and classifier-free guidance during diffusion model sampling. As per the ``score matching'' interpretation of diffusion models, we assume that our trained noise network approximates the score function of the true conditional latent distribution $p(z|e_i)$ as $\epsilon_\psi(z^t,t,e_i) \approx -\sigma_t \nabla_{z^t} \log p(z^t|e_i)$. For classifier guidance, we can perturb our diffusion sampling by adding the gradient of the log likelihood of our CLIP encoder $p_\psi(e_i|z_t)$ to the diffusion score as follows
\begin{equation}
    \tilde{\epsilon}_\psi(z^t,t,e_i) = \epsilon_\psi(z^t,t,e_i) - \eta \sigma_t \nabla_{z^t} \log p_\psi(e_i|z^t) \approx -\sigma_t \nabla_{z^t} \left[  \log p(z^t|e_i) + \eta \log p_\psi(e_i|z^t) \right] \text{.}
\end{equation}

We can rewrite this as classifier guidance on the unconditional score $\nabla_{z^t}\log p(z^t)$ with
\begin{equation}\label{eq-diffusion-classguidance}
    -\sigma_t \nabla_{z^t} \left[  \log p(z^t) + \gamma \log p(e_i|z^t) \right] \quad \text{ with } \quad \gamma = 1 + \eta
\end{equation}
using Bayes' rule, as $\log p(z^t|e_i) = \log p(e_i|z^t) + \log p(z^t) - \log p(e_i)$, and thus $\nabla_{z^t}\log p(z^t|e_i) = \nabla_{z^t}\log p(e_i|z^t) + \nabla_{z^t}\log p(z^t)$. 

For classifier-free guidance, we aim to perform the above sampling without access to a classifier, as long we possess a conditional diffusion model $\epsilon_\psi(z^t,t,e_i)$ that doubles as an unconditional model $\epsilon_\psi(z^t,t,0)$, as illustrated in Section \ref{section:class-free}.

Using Bayes' rule again, we can see that $\nabla_{z^t}\log p(e_i|z^t) = \nabla_{z^t}\log p(z^t|e_i) - \nabla_{z^t}\log p(z^t)$. If we substitute this into Eq.~\ref{eq-diffusion-classguidance} we obtain
\begin{align}
    & -\sigma_t \nabla_{z^t} \left[  \log p(z^t) + \gamma \left( \log p(z^t|e_i) - \log p(z^t) \right) \right] \text{,}\\
    & -\sigma_t \nabla_{z^t} \left[ \left( 1 - \gamma \right) \log p(z^t) + \gamma \log p(z^t|e_i) \right] \text{,}
\end{align}
which can be implemented with our conditional network as
\begin{equation}
    \tilde{\epsilon}_\psi(z^t,t,e_i) = \left(1 - \gamma\right) \epsilon_\psi(z^t,t,0) + \gamma \epsilon_\psi(z^t,t,e_i) \text{.}
\end{equation}

\subsection{Network architectures} \label{appendix:architecture}

\paragraph{Base Network ($f$)} Our choice for a base model is a CLIP-Adapter \citep{gao_clip-adapter_2021}, which consists of a frozen CLIP image encoder \citep{radford_learning_2021} with added learned fully-connected layers refining the output embedding. Specifically, we use the \textit{ViT-L/14@336px} CLIP encoder type with embedding size of 768, and for the adapter we use a MLP with one hidden layer of 256 units, which are followed by a rectified linear activation, and a new linear output layer of size 768. The advantages of this model choice lie in its combination of high base performance (due to pre-trained knowledge contained in the CLIP component) and relatively small parameter count, enabling agile medium-small scale experiments. This base CLIP-Adapter network purely works as a base model and is not to be confused with HyperCLIP, which is employed at the meta-level. In Section \ref{sec:experiment-results}, when benchmarking the base model alone in the zero-shot setting (CLIP as Base Model), we drop the Adapter and use pre-trained zero-shot CLIP.

\paragraph{Hypernetwork ($h$)} For the hypernetworks used in our baseline as well as as the generative model, we use a MLP with one hidden layer of 256 units, which are followed by a rectified linear activation. For the unconditioned hypernetwork, the embedding to the hypernetwork is a vector of dimension 64, while for the conditioned counterpart, the task embedding is used. In order to ensure that the generated weights are properly normalized at initialization, we use the Kaiming initialization \citep{he_delving_2015} for the hypernetwork weights, initialize the embedding as a sample from a multivariate standard gaussian distribution (for unconditioned models), and use the NTK parametrization \citep{jacot_neural_2020} for the target network.

\paragraph{Variational Autoencoder} For the variational autoencoder used as our unconditioned generative model, we use an MLP of 2 hidden layers of size 512 and 256, each followed by the rectified linear non-linearity. We chose 32 as the latent code dimension. We use the same architecture for the decoder, except for the dimensionality of the 2 hidden layers being swapped. We use the Kaiming initialization \citep{he_delving_2015} to initialize the weight of both the encoder and decoder.

\paragraph{HyperCLIP} We parametrize our HyperCLIP model as a fully-connected MLP with a single hidden layer of dimension 256, taking as input the flattened weight of the base network and outputting the corresponding CLIP encoding. We chose the tangent hyperbolic function as the activation function in the hidden layer.

\paragraph{HyperLDM} While the original LDM makes use of a time-conditional UNet \citep{ronneberger_u-net_2015} to parametrize the noise network, we are unfortunately unable to make use of spatial information and convolutions due to the non-spatial nature of our latent space. We parametrize our HyperLDM as a fully-connected network with residual connections and squeeze-and-excitation layers \citep{hu_squeeze-and-excitation_2018}. The time index $t$ is embedded into a vector with a 150-dimensional sinusoidal positional embedding, and is concatenated together with the task-conditional embedding $e_i$ at the input layer and at intermediate activations. Hidden layer dimensions are 8192, 16384, 8192.

\subsection{Notes and Limitations}
While optimal parameter counts for the task-conditional techniques vary between our techniques, as well as between our techniques and the baselines, all of the investigated approaches ultimately produce adapted weights for the same base network $f$, with the same architecture. The average performance of this base network with this fixed architecture when adapted and deployed on each individual test task is what allows us to fairly compare all meta-learning and multi-task algorithms. We acknowledge that, as a limitation of our work, the comparisons hold up when comparing relatively small fixed base networks $f$, and our approach might not be scalable to compete with massively pre-trained large scale multi-task models. In any case, we believe that weight space generation of compact models can be useful in a variety of contexts, such as when the adapted base model needs to be deployed in embedded systems and other domains with limited compute resources. The scaling behavior of our techniques is still an open problem, which can be of interest for future research.

\subsection{The Meta-VQA Dataset} \label{appendix:meta-vqa}
The original VQA problem is about choosing a suitable natural language answer $y_k$ when prompted with both a natural language question $t_k$ and an image $x_k$. Our observation is that the VQA problem can then easily be reformulated as a meta-learning image classification problem with natural language task descriptions: given question-image-answer triples $(t_k,x_k,y_k) \in D$, we can group the data by unique questions $t_i$ (which will serve as task descriptor), each of which can then be associated with supervised image classification tuples $(x_j^i,y_j^i) \in D_i$. To make sure the designed tasks are meaningful, we filter out question-answer pairs with questions in choosing form, \textit{e.g.,}~“A or B?” or yes/no answers. From the remaining questions we keep the ones which appear at least 20 times throughout the dataset, such that each task contains enough samples. In the end our Meta-VQA dataset is composed of 1234 unique tasks (questions), split into 870 training tasks and 373 test tasks, for a total of 104112 image-answer pairs. There are on average 9.13 answer choices per question/task. The average size of the support set is 57.85 examples, while the average size of the query set is 25.9 examples.

\subsection{Baseline methods} \label{appendix:method-details}

We detail an overview of the baseline methods we benchmark in table \ref{tab:method_details}, together with algorithm tables detailing each baseline method.

\paragraph{Training:} The number of epochs each model is trained on, the learning rate $\texttt{lr}$ of the optimization, as well as the learning rate and number of steps of the adaptation algorithm used for each method can be found in table \ref{tab:method_hp}. For all methods using an adaptation $\mathcal{A}_{\mathcal{T}_i}$, the dataset $D_i$ from a sampled task $\mathcal{T}_i$ is randomly split into a support set $D_i^{\text{s}}$ and a query set $D_i^{\text{q}}$ during training. The support set is then used to perform the adaptation, while the query set is used to compute the loss on which the meta-parameters are updated (see Section \ref{sec:maml}). For baselines with no inner-loop adaptation $\mathcal{A}_{\mathcal{T}_i}$, all the data $D_i$ from a sampled task $\mathcal{T}_i$ is used in training. \textit{Unconditional} methods do not have access to the task embedding $e_i$, while \textit{conditional} methods do. When the percentage of available task descriptors is reduced, conditional baselines are trained only on the tasks for which the descriptors are available, as they require such descriptor during training, unlike our two-phased techniques.

\paragraph{Evaluation:} For each held-out test task $\mathcal{T}_i$ from the Meta-VQA dataset, we perform zero-shot model evaluation on the fixed predefined query set $D_i^{\text{q}}$ for the task. Zero-shot performance is evaluated before applying any adaptation procedure $\mathcal{A}_{\mathcal{T}_i}$.

\begin{algorithm}[h]
\caption{Unconditional Multitask Training}\label{algo:uncond-multitask}
\begin{algorithmic}
\State Define the base network $f$ with parameters $W$.
\For{$\text{epoch}=1,\ldots,N$}
    \State Sample a training batch of image-answer pairs $(x_k,y_k)$ from a mix of random training tasks $\mathcal{T}_{i}$.
    \State Update $W$ with gradient descent computed with respect to the classification loss over the sampled batch.
\EndFor
\end{algorithmic}
\end{algorithm}

\begin{algorithm}[H]
\caption{Unconditional MNet-MAML Training}\label{algo:uncond-mnet-maml}
\begin{algorithmic}
\State Define the base network $f$ with parameters $W$.
\For{$\text{meta-epoch}=1,\ldots,N$}
    \State Sample a training task $\mathcal{T}_{i}$ and data $D_i$.
    \State Randomly split $D_i$ into support set $D_i^s$ and query set $D_i^q$.
	\State Run inner-loop adaptation $\mathcal{A}_{\mathcal{T}_i}$ using the support set $D_i^s$, fine-tuning $W$ into task-adapted $W_i = \mathcal{A}_{\mathcal{T}_i}(W)$.
	\State Use MAML gradient update to adapt $W$ given the inner-loop adaptation.
\EndFor
\end{algorithmic}
\end{algorithm}

\begin{algorithm}[H]
\caption{Unconditional HNet-MAML Training}\label{algo:uncond-hnet-maml}
\begin{algorithmic}
\State Define the base network $f$ with parameters $W$.
\State Define a hypernetwork $h$ with meta-parameters $\theta$, mapping a latent vector $z^0$ to base network weights $W$.
\For{$\text{meta-epoch}=1,\ldots,N$}
    \State Sample a training task $\mathcal{T}_{i}$ and data $D_i$.
    \State Randomly split $D_i$ into support set $D_i^s$ and query set $D_i^q$.
	\State Run inner-loop adaptation $\mathcal{A}_{\mathcal{T}_i}$ using the support set $D_i^s$, fine-tuning $z^0$ into task-adapted $z_i = \mathcal{A}_{\mathcal{T}_i}(z^0)$.
	\State Use MAML gradient update to adapt $z^0$ and $\theta$ given the inner-loop adaptation.
\EndFor
\end{algorithmic}
\end{algorithm}

\begin{algorithm}[h]
\caption{Conditional Multitask Training}\label{algo:cond-multitask}
\begin{algorithmic}
\State Define the base network $f$ with parameters $W$.
\State Define a hypernetwork $h$ with meta-parameters $\theta$, mapping the clip embedding $e_i$ of the language task descriptor to base network weights $W_i$.
\For{$\text{epoch}=1,\ldots,N$}
    \State Sample a training batch of task clip embedding, image and answer triples $(e_k,x_k,y_k)$ from a mix of random training tasks $\mathcal{T}_{i}$.
    \State Update $\theta$ with gradient descent computed with respect to the classification loss over the sampled batch.
\EndFor
\end{algorithmic}
\end{algorithm}

\begin{algorithm}[h]
\caption{Conditional Multitask FiLM Training}\label{algo:cond-multitask}
\begin{algorithmic}
\State Define the base network $f$ with parameters $W$.
\State Define a FiLM layer, mapping the clip embedding $e_i$ of the language task descriptor to modulation signals for the hidden activation layer of $f$.
\For{$\text{meta-epoch}=1,\ldots,N$}
    \State Sample a training batch of task clip embedding, image and answer triples $(e_k,x_k,y_k)$ from a mix of random training tasks $\mathcal{T}_{i}$.
    \State Update $\theta$ with gradient descent computed with respect to the classification loss over the sampled batch.
\EndFor
\end{algorithmic}
\end{algorithm}

\begin{algorithm}[H]
\caption{Conditional HNet-MAML Training}\label{algo:uncond-hnet-maml}
\begin{algorithmic}
\State Define the base network $f$ with parameters $W$.
\State Define a hypernetwork $h$ with meta-parameters $\theta$, mapping the clip embedding $e_i$ of the language task descriptor to base network weights $W_i$.
\For{$\text{meta-epoch}=1,\ldots,N$}
    \State Sample a training task $\mathcal{T}_{i}$, data $D_i$ and the clip embedding $e_i$ of the task descriptor.
    \State Randomly split $D_i$ into support set $D_i^s$ and query set $D_i^q$.
	\State Run inner-loop adaptation $\mathcal{A}_{\mathcal{T}_i}$ using the support set $D_i^s$, fine-tuning $e_i$ into task-adapted $\tilde{e_i} = \mathcal{A}_{\mathcal{T}_i}(e_i)$.
	\State Use MAML gradient update to adapt $\theta$ given the inner-loop adaptation.
\EndFor
\end{algorithmic}
\end{algorithm}

\begin{table}[H]
\caption{Overview of the different methods trained on MetaVQA. The \textbf{parameters} are optimized via the task loss evaluated on the output of the \textbf{function}, averaged over minibatches of tasks. The adaptation $\mathcal{A}_{\mathcal{T}_i}$ implements a few step gradient descent algorithm applied on the argument parameter, w.r.t the task loss evaluated on the support set. }
\label{tab:method_details}
\begin{center}
\begin{tabular}{l l l}
\toprule
\multicolumn{1}{c}{\bf Method}  & \multicolumn{1}{c}{\bf Function} & \multicolumn{1}{c}{\bf Parameters} 
\\ \midrule
Uncond. Multitask   & $f(\cdot, W)$  & $W$ \\
Uncond. MNet-(FO)MAML & $f(\cdot, \mathcal{A}_{\mathcal{T}_i}(W^0))$ & $W^0$ \\
Uncond. HNet-MAML & $f(\cdot, h(\mathcal{A}_{\mathcal{T}_i}(z^0),\theta))$ &  $\theta, z^0$ \\
\midrule
Cond. Multitask     & $f(\cdot, h(e_i,\theta))$& $\theta$ \\
Cond. Multitask FiLM    & $f(\cdot, e_i, W)$& $W$ \\
Cond. HNet-MAML & $f(\cdot, h(\mathcal{A}_{\mathcal{T}_i}(e_i),\theta))$ & $\theta$ \\
\bottomrule
\end{tabular}
\end{center}
\end{table}

\begin{table}[H]
\caption{Hyperparameters used for the baseline methods. All methods are trained with the Adam \citep{kingma_adam_2017} optimizer, with meta-batch size of 32 tasks. We use gradient norm clipping for all optimization, with the maximum norm set to 10. Note that when the adaptation algorithm $\mathcal{A}$ has a range of possible steps, the number of step is sampled uniformly from the range for every adaptation.}
\label{tab:method_hp}
\begin{center}
\begin{tabular}{l l l l l l}
\toprule
\multicolumn{1}{c}{\bf Method}  & \texttt{epochs} &\texttt{lr}&$\mathcal{A}$\texttt{-lr} & $\mathcal{A}$\texttt{-steps} 
\\ \midrule
Uncond. MNet-Multitask   & 300  & 0.0001 & - & -  \\
Uncond. MNet-(FO)MAML  &  500  & 0.00003 & 0.01 & 0-10  \\
Uncond. HNet-MAML & 100  & 0.00003 & 0.1 & 0-10  \\
\midrule
Cond. Multitask     &  60  & 0.0001 & - & -  \\
Cond. Multitask FiLM & 300 & 0.0001 & - & - \\
Cond. HNet-MAML & 200  & 0.00001 & 0.1 & 0-10  \\
\bottomrule
\end{tabular}
\end{center}
\end{table}

\subsection{Guidance Models} \label{appendix:our-method-details}

\subsubsection{Generative hypernetwork} \label{appendix:generative-hypernet}

To enable our guidance methods, we need to first train a generative hypernetwork $h$ as in Section \ref{sec:hypernet-generative}, either in the form of an Unconditional Hypernetwork, or of a Hypernetwork VAE:
\begin{itemize}
\item For \textbf{HNet + HyperCLIP guidance} and \textbf{HNet + HyperLDM}, we meta-learnt an unconditional hypernetwork with the exact same hyperparameters as the baseline \textbf{Uncond. HNet-MAML}, and used it as the generative hypernetwork. 
\item For \textbf{HVAE + HyperCLIP guidance} and \textbf{HVAE + HyperLDM}, we trained a VAE on samples of fine tuned network weights $W_i$ using the base network architecture specified in Appendix \ref{appendix:architecture}. We detail the procedure in Algorithm \ref{algo:hvae} and, as training samples $W_i$, we use adaptations over the base network (initialized from a learned \textbf{Uncond. MNet-MAML} initialization), using 50-step adaptation $\mathcal{A}_{\mathcal{T}_i}$ with learning rate $0.01$ on randomly split support sets. We trained the VAE for 2000 epochs where each epoch is a single pass through all the tasks, with the Adam \citep{kingma_adam_2017} optimizer and $0.0001$ learning rate and batch size 32. We used gradient norm clipping independently for both the encoder and decoder, with the maximum norm capped at 1000.
\end{itemize}


\begin{algorithm}[H]
\caption{HVAE Training}\label{algo:hvae}
\begin{algorithmic}
\State Define the base network $f$ with parameters $W$.
\State Define an encoder $z = d(W,\omega)$ with parameters $\omega$ and a hypernetwork decoder $W = h(z,\theta)$ with parameters $\theta$.
\State Obtain a previously learned base network initialization $W^0$ according to \textbf{Uncond.~MNet-MAML} (Algorithm \ref{algo:uncond-hnet-maml}).
\For{$\text{epoch}=1,\ldots,N$}
    \State Create an empty batch $B = \lbrace\rbrace$.
    \For{$b = 1,\ldots,M$}
        \State Sample a training task $\mathcal{T}_{i}$ and data $D_i$.
        \State Randomly split $D_i$ into support set $D_i^s$ and query set $D_i^q$.
    	\State Run inner-loop adaptation $\mathcal{A}_{\mathcal{T}_i}$ using the support set $D_i^s$, fine-tuning $W_i = \mathcal{A}_{\mathcal{T}_i}(W^0)$.
    	\State Add the fine-tuned weights to the batch: $B = B \cup \lbrace W_i\rbrace$.
	\EndFor
	\State Train the HVAE encoder and decoder using the VAE loss to reconstruct the weight batch $B$.
\EndFor
\end{algorithmic}
\end{algorithm}

\subsubsection{HyperCLIP}

\paragraph{Training} In order to train the HyperCLIP model, we need samples of fine tuned network weights $W_i$. We use adaptations from \textbf{Uncond. HNet-MAML}, using 50-step adaptation $\mathcal{A}_{\mathcal{T}_i}$ with learning rate $0.1$, on randomly split support sets. We trained our HyperCLIP model for 600 epochs with the Adam \citep{kingma_adam_2017} optimizer, $0.0003$ learning rate and batch size 64 for all our experiments.

\paragraph{Guidance} We use 10 steps guidance with $\lambda=0.01$ and learning rate $0.1$, to perform guidance within either the HNet or HVAE latent spaces.

\paragraph{Evaluation} For each held-out test task $\mathcal{T}_i$ from the Meta-VQA dataset, we perform zero-shot model evaluation on the fixed predefined query set $D_i^{\text{q}}$ for the task. Zero-shot performance is evaluated on the output of the generative hypernetwork $h$ after applying latent space guidance.

\begin{algorithm}[H]
\caption{HNet + HyperCLIP Training}\label{algo:hnet-hyperclip}
\begin{algorithmic}
\State Learn an unconditional hypernetwork $h(z^0,\theta)$ with the \textbf{Uncond. HNet-MAML} procedure from Algorithm \ref{algo:uncond-hnet-maml}.
\State Learn HyperCLIP network $\text{CLIP}_H(W)$ using the HyperCLIP training procedure from Algorithm \ref{algo:hyperclip}. For sampling fine-tuned $W_i$, fine-tune the base-network on training tasks.
\end{algorithmic}
\end{algorithm}

\begin{algorithm}[H]
\caption{HVAE + HyperCLIP Training}\label{algo:hnet-hyperclip}
\begin{algorithmic}
\State Learn an unconditional hypernetwork $h(z,\theta)$, as the decoder of a HVAE (Algorithm \ref{algo:hvae}).
\State Learn HyperCLIP network $\text{CLIP}_H(W)$ using the HyperCLIP training procedure from Algorithm \ref{algo:hyperclip}. For sampling fine-tuned $W_i$, fine-tune the base-network on training tasks.
\end{algorithmic}
\end{algorithm}

\begin{algorithm}[H]
\caption{HyperCLIP Guidance (Inference time)}\label{algo:hyperclip-guidance-inference}
\begin{algorithmic}
\State Define a learned unconditional hypernetwork $h(z,\theta)$, as either a HNet $h(z^0,\theta)$ (Algorithm \ref{algo:uncond-hnet-maml}) or the decoder of a HVAE (Algorithm \ref{algo:hvae}).
\State Define a learned HyperCLIP network $\text{CLIP}_H(W)$.
\State Define an unseen task $\mathcal{T}_i$ with natural language task descriptor $t_i$.
\State Randomly sample $z\sim \mathcal{N}(0,I)$ if using the decoder of a HVAE, or set $z=z^0$ where $z^0$ is the meta learned embedding initialization of the Hnet.
\State Optimize $z$ with gradient descent over $\mathcal{L}_\text{guidance}(z)$ (Eq.~\ref{eq:hyperclip-guidance}), obtaining \textit{guided} $z_i$.
\State Obtain \textit{guided} base weights $W_i = h(z_i, \theta)$.
\State Use adapted base network $f$ with weights $W_i$ to classify examples from the unseen task $\mathcal{T}_i$.
\end{algorithmic}
\end{algorithm}

\subsubsection{HyperLDM}

\paragraph{Training} Similarly to HyperCLIP, to train HyperLDM we need samples of fine tuned network weights $W_i$, for which we use adaptations from \textbf{Uncond. HNet-MAML}, using 50-step adaptation $\mathcal{A}_{\mathcal{T}_i}$ with learning rate $0.1$, on a randomly split support sets. We parametrize the diffusion process with a linear noise schedule, $\beta$ starting at 0.0001 and ending at 0.06, and 350 diffusion timesteps. We train the HyperLDM for 1000 epochs with the Adam optimizer, $0.00025$ learning rate and 128 epochs, for all our experiments.

\paragraph{Evaluation} Evaluation is performed as for HyperCLIP guidance, except for the fact that adaptation is performed natively through sampling from the learned reversed diffusion process, with parameters derived from the chosen $\beta$ schedule. The guidance parameter $\gamma>0$ can be tuned during inference to accentuate the effect of classifier-free guidance.

\begin{algorithm}[H]
\caption{HNet + HyperLDM Training}\label{algo:hnet-hyperldm}
\begin{algorithmic}
\State Learn an unconditional hypernetwork $h(z^0,\theta)$ with the \textbf{Uncond. HNet-MAML} procedure from Algorithm \ref{algo:uncond-hnet-maml}.
\State Learn the HyperLDM network $\epsilon_\psi(z^t,t,e_i)$ using the HyperLDM training procedure, optimizing reconstruction of $z_i^0$ with loss from Eq.~\ref{eq:hyperldm-loss}. For sampling fine-tuned $z_i$, fine-tune the base-network on training tasks, then encode the weights using the HNet.
\end{algorithmic}
\end{algorithm}

\begin{algorithm}[H]
\caption{HVAE + HyperLDM Training}\label{algo:hnet-hyperldm}
\begin{algorithmic}
\State Learn an unconditional hypernetwork $h(z,\theta)$, as the decoder of a HVAE (Algorithm \ref{algo:hvae}).
\State Learn the HyperLDM network $\epsilon_\psi(z^t,t,e_i)$ using the HyperLDM training procedure, optimizing reconstruction of $z_i^0$ with loss from Eq.~\ref{eq:hyperldm-loss}. For sampling fine-tuned $z_i$, fine-tune the base-network on training tasks, then encode the weights using the HVAE.
\end{algorithmic}
\end{algorithm}

\begin{algorithm}[H]
\caption{HyperLDM Inference}\label{algo:hyperldm-inference}
\begin{algorithmic}
\State Define a learned unconditional hypernetwork $h(z,\theta)$, as either a HNet $h(z^0,\theta)$ (Algorithm \ref{algo:uncond-hnet-maml}) or the decoder of a HVAE (Algorithm \ref{algo:hvae}).
\State Define a learned HyperLDM network $\epsilon_\psi(z^t,t,e_i)$.
\State Define an unseen task $\mathcal{T}_i$ with natural language task descriptor $t_i$, with clip embedding $e_i$.
\State Randomly sample $z\sim \mathcal{N}(0,I)$.
\State Iteratively modify $z$ with diffusion sampling using the learned $\epsilon_\psi$ network, obtaining \textit{guided} $z_i$.
\State Obtain \textit{guided} base weights $W_i = h(z_i, \theta)$.
\State Use adapted base network $f$ with weights $W_i$ to classify examples from the unseen task $\mathcal{T}_i$.
\end{algorithmic}
\end{algorithm}

\subsection{Few-Shot Learning}
\label{appendix:fewshot}

For completeness, we include in Table \ref{tab:fshot} the results for few-shot learning on the test split of Meta-VQA. Our technique, unlike classic MAML, does not optimize specifically for the few-shot learning setting. Instead, the few-shot learning results are meant to contextualize performance gains in the zero-shot setting: zero-shot performance gains should be interpreted as relative to the few-shot accuracy ceiling of 60.24\%, the maximum attained with our fixed choice of base model.

For few-shot learning at test time, all adaptation is performed on the support set of the test tasks. For MAML baselines, we keep the same adaptation-time learning rate as during training, and we always adapt for 50 steps. For each multitask baseline, we use the same adaptation scheme (steps, learning rate, adapting parameters) as their MAML counterpart.

\begin{table}[ht]
\caption{Few-Shot learning accuracy averaged over Meta-VQA test tasks. (* ours)}
\label{tab:fshot}
\begin{center}
\begin{tabular}{l l}
\toprule
\multicolumn{1}{c}{\bf Method}  & \multicolumn{1}{c}{\bf Few-Shot}
\\ \midrule
CLIP as Base Model & 54.93 ($\pm$ 0.11) \\
\midrule
Uncond.~Multitask   & 55.53 ($\pm$ 0.40) \\
Uncond.~MNet-MAML   &  \textbf{60.24} ($\pm$ 0.84) \\
Uncond.~MNet-FOMAML &  60.03 ($\pm$ 0.48) \\
Uncond.~HNet-MAML   &  58.70 ($\pm$ 0.10) \\
\midrule
Cond.~Multitask     & 59.46 ($\pm$ 0.31) \\
Cond.~HNet-MAML     & 59.48 ($\pm$ 0.03) \\
\bottomrule
\end{tabular}
\end{center}
\end{table}

\end{document}